\documentclass[lettersize,onecolumn]{IEEEtran}
\usepackage{amsmath,amsfonts}
\usepackage{algorithmic}
\usepackage{algorithm}
\usepackage{array}
\usepackage[caption=false,font=normalsize,labelfont=sf,textfont=sf]{subfig}
\usepackage{textcomp}
\usepackage{stfloats}
\usepackage{url}
\usepackage{caption}
\usepackage{subcaption}
\usepackage{verbatim}
\usepackage{graphicx}
\usepackage{cite}
\usepackage{algorithm}
\usepackage{algorithmic}

\usepackage{amsfonts,amsmath,amsthm, mathtools}
\usepackage{amssymb}
\usepackage{caption}
\usepackage{verbatim}
\usepackage{mathrsfs}
\usepackage{algorithm}
\usepackage{algorithmic}
\usepackage{comment}
\usepackage{multirow, tabularx, booktabs}
\usepackage[dvipsnames]{xcolor}

\usepackage{hyperref}
\usepackage{dsfont}

\newtheorem{theorem}{Theorem}

\newtheorem{lemma}[theorem]{Lemma}

\newtheorem{definition}[theorem]{Definition}

\newtheorem{remark}[theorem]{Remark}

\usepackage{fullpage}
\usepackage{xspace}

\numberwithin{equation}{section}

\theoremstyle{plain}

\newtheorem*{theorem*}{Theorem}

\theoremstyle{definition}

\usepackage[margin=1in]{geometry}

\newcommand{\rr}{\mathbb{R}}

\newcommand{\zc}{\mathcal{z}}

\newcommand{\mm}{\mathcal{M}}

\newcommand{\xl}{\mathcal{X}}
\newcommand{\yl}{\mathcal{Y}}
\newcommand{\ec}{\mathcal{E}}
\newcommand{\ee}{\mathbb{E}}

\newcommand{\tn}{\mathbb{T}}

\newcommand{\dd}{\mathbb{D}_c}

\newcommand{\xx}{\mathcal{X}}

\begin{document}

\title{On the Universal Statistical Consistency of Expansive Hyperbolic Deep Convolutional Neural Networks}

\author{Sagar Ghosh, Kushal Bose, and Swagatam Das}



\maketitle

\begin{abstract}
The emergence of Deep Convolutional Neural Networks (DCNNs) has been a pervasive tool for accomplishing widespread applications in computer vision. Despite its potential capability to capture intricate patterns inside the data, the underlying embedding space remains Euclidean and primarily pursues contractive convolution. Several instances can serve as a precedent for the exacerbating performance of DCNNs. The recent advancement of neural networks in the hyperbolic spaces gained traction, incentivizing the development of convolutional deep neural networks in the hyperbolic space. In this work, we propose Hyperbolic DCNN based on the Poincar\'{e} Disc. The work predominantly revolves around analyzing the nature of expansive convolution in the context of the non-Euclidean domain. We further offer extensive theoretical insights pertaining to the universal consistency of the expansive convolution in the hyperbolic space. Several simulations were performed not only on the synthetic datasets but also on some real-world datasets. The experimental results reveal that the hyperbolic convolutional architecture outperforms the Euclidean ones by a commendable margin. 
\end{abstract}


\section{Introduction}
The ubiquitous utility of Deep Convolutional Neural Networks (DCNNs) \cite{cnn} dominated the arena of Computer Vision \cite{image, rs, car} over the past decade. This profound success can be attributed to the effectiveness of the CNNs in approximating the broader class of continuous functions \cite{lin}. The prevalent convolutional neural architectures \cite{resnet, vggnet} predominantly operate in the Euclidean feature space. The choice of Euclidean space is mostly for implementable closed-form vector space and inner product structures and their availability in tabular forms. We are focussed on DCNN architectures which evolve around $1-$ dimensional convolution based on one input channel and ReLU (Rectified Linear Unit, $r(x):=\max(0,x)$ for $x\in\rr$) activation function given to the computational units (Neurons). For two functions $f,g:\rr^n\to\rr$, we define their convolution as 
\begin{align*}
    f\otimes g(z):=\int_{\rr^n}f(x)g(z-x)dx,
\end{align*}
where $z\in\rr^n$. In the discrete version, given a filter $\bf{w}:=\{w_i\}_{i=-\infty}^\infty$, where only finitely many $w_j\neq 0$. We call $\bf{w}$ to be a filter of length $s$ if $w_j\neq 0$ only for $0\leq j\leq s$. For a one dimensional input vector $\bf{v}:=\{v_1,v_2,...,v_n\}\in\rr^n$, we can define two types of convolution operations, namely \textit{Expansive Convolution} ($\bf{w}\ast\bf{v}$) and \textit{Contractive Convolution}($\bf{w}\star\bf{v}$), given by the following forms of equations
\begin{equation}\label{eqn:i1}
    (\textbf{w}\ast\textbf{v})_k:=\sum_{i=1}^nw_{k-i}v_i, \hspace{1ex} k=1,2,...,n+s
\end{equation}
and 
\begin{equation}\label{eqn:i2}
    (\textbf{w}\star\textbf{v})_k:=\sum_{i=k-s}^kw_{k-i}v_i, \hspace{1ex} k=s+1,s+2,...,n
\end{equation}
respectively. Now for a set of $L$ filters $\{\bf{w}_i\}_{i=1}^L$, where $L$ is the depth of our network with $L$ many bias vectors $\{b_i\}_{i=1}^L$, we recursively define the output of an intermediate layer given in terms of the output of the previous layer as \cite{har}
\begin{align*}
    h_i(x)=r\left(\textbf{w}_i\circ h_{i-1}(x)+b_i\right), \hspace{1ex} \text{for}\hspace{1ex} i=1,2,...,L,
\end{align*}
starting with the input as $h_0(x)=x$ and $\circ$ can be either $\ast$ or $\star$ as defined by equations \ref{eqn:i1} or \ref{eqn:i2} respectively. The final one-dimensional output of this network is defined as the scalar product between the output produced by the $L^{th}$ layer with a trainable vector $a_L$ of compatible length
\begin{align*}
    h_o(x):=a_l\cdot h_L(x).
\end{align*}

Although this form of Euclidean convolution has been proven to be enormously successful in several tasks in computer vision, several precedents \cite{ex1, ex2, ex3} can be put forth where Euclidean feature space seems unproductive, like datasets containing hierarchical structures.

The embedding learning of hierarchical datasets in the Euclidean spaces raises concerns regarding extracting meaningful information. To address this concern, the research community showed numerous endeavors \cite{gan, poincare_image, bde} in designing neural networks in non-Euclidean feature spaces. Hyperbolic Neural Networks (HNNs) \cite{gan} paved the way for generating embeddings in the hyperbolic space equipped with negative curvature. HNNs offer congenial feature space to understand the complex relationships and structural intricacies within the data. In this sobering context, the Hyperbolic Deep Convolutional Neural Networks (HDCNN) \cite{bde} successfully pursued various image-related tasks. Despite the profound impact of hyperbolic neural networks, insightful theoretical analysis is lacking in the literature. This work is motivated to bridge this widened gap by offering an extensive statistical analysis to comprehend the underlying working mechanism of hyperbolic convolution.     
The consistency analysis of Euclidean convolutional networks has been addressed in \cite{lin}, where the authors provided theoretical insights through explicit bounds on packing numbers and error analysis for bounded samples within the Euclidean framework. Their work primarily focuses on expansive convolutional operations. However, a similar analysis for convolution operations in hyperbolic space remains uncharted. In particular, exploring the consistency of expansive convolution in hyperbolic spaces would mark a significant advancement toward a deeper theoretical understanding of hyperbolic convolution.

In response to this gap, we have developed a theoretical foundation for the consistency of 1-D expansive Deep Hyperbolic Deep Convolutional Neural Networks (eHDCNNs). Our proposed architecture offers a hyperbolic generalization of the standard Euclidean DCNN, which is essential for defining statistical concepts and enabling subsequent theoretical analysis. We initially constructed a framework for 1-D expansive convolution on the Poincaré disc, providing a basis for our consistency analysis. Additionally, experiments conducted on synthetic and real-world datasets illustrate the efficiency of representing features in hyperbolic spaces. The results empirically support our theoretical findings, achieving lower error rates in the hyperbolic setting significantly faster than the Euclidean equivalent.

\vspace{5pt}
\noindent
\textbf{Contribution} Our main contributions could be summarized in the following way:
\begin{itemize}
\item We provide theoretical insights, including the consistency analysis of the expansive $1$-D convolution in hyperbolic space. To the best of our knowledge, this is the first work to present a complete proof in the context of a fully hyperbolic set-up. In doing so, we have also introduced the concept of a fully hyperbolic convolution operation on the Poincar\'{e} Disc, which is the generalization of the conventional Euclidean convolution operation on hyperbolic spaces. Additionally, we have defined several statistical terminologies within the hyperbolic framework to derive universal consistency. All proofs and derivations are provided in the Appendix.

\item Our experimental simulations demonstrate that eHDCNN training converges more rapidly than the training of the Euclidean DCNN, which we have already established theoretically. The faster reduction of error rate reaffirms the requirements of the lower number of training iterations for hyperbolic convolutional networks compared to their conventional Euclidean counterparts, establishing the effectiveness of eHDCNN.  
\end{itemize}

\begin{figure}[!ht]
    \centering
     \includegraphics[width=0.8\textwidth]{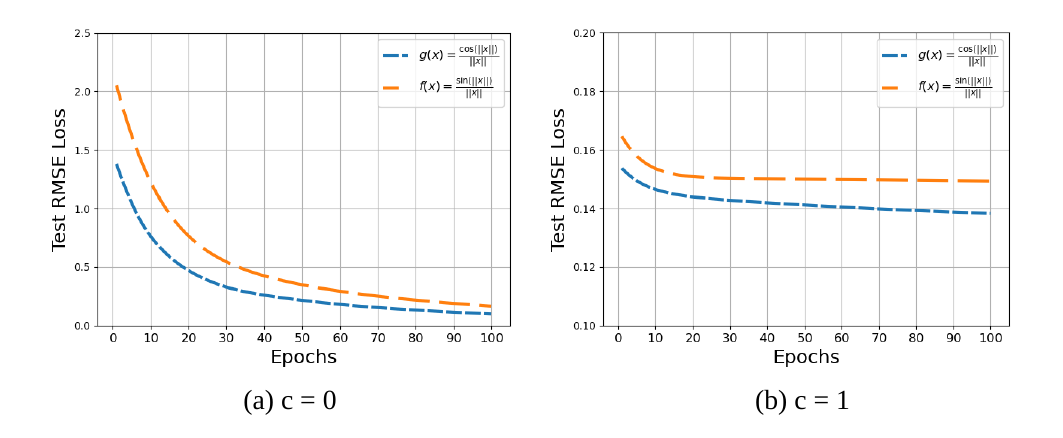}     
     \caption{Test Root Mean Squared error for $f(x)$ and $g(x)$ plotted using (a) eDCNN architecture curvature $0$ (i.e., Euclidean space) and (b) eHDCNN architecture with curvature $1$. }
     \label{fig:motivating}
\end{figure}

\section{A Motivating Example}
We will demonstrate the efficacy of our proposed hyperbolic expansive convolution over conventional Euclidean expansive convolution through the following simulation. 

\vspace{5pt}
\noindent
\textbf{Experimental Setup} Consider the following functions,
\begin{align*}
    f(x)=\frac{\sin(\|x\|_2)}{\|x\|_2},  \hspace{3ex} g(x)=\frac{\cos(\|x\|_2)}{\|x\|_2},
\end{align*}
where $f(x)$ and $g(x)$ both are modeled as regression task like $y:=h(x)+\epsilon$. Here, $h$ can be replaced with either $f$ or $g$. The training instances are generated by sampling $\epsilon\sim \mathcal{N}(0,0.01)$ and $x\sim unif([-1,1]^{5})$. A total of $1000$ instances will be generated for both cases where $800$ and $200$ samples will be respectively used for training and testing. Importantly, the test samples are considered without the Gaussian noise. The filter length is fixed at $8$ with the number of layers is $4$. Both models are trained for $100$ epochs over the training set. The test Root Mean Squared Loss (RMSE) is recorded after the completion of training and presented in Figure \ref{fig:motivating}. Experiments are conducted for two different curvatures $c=0$ (Euclidean space) and $c=1$. We considered unit radius Poincar\'{e} Disc as the hyperbolic space. Assuming the point set in discrete metric space, we employed Gromov Hyperbolicity (GH) \cite{gromov} to measure the hyperbolicity ($\delta$) of the corresponding data points. The metric offers hyperbolicity of $f$ and $g$ are respectively $\delta_f = 0.45$ and $\delta_g = 0.13$, indicating that $g$ is more hyperbolic comparing to $f$.  

\vspace{5pt}
\noindent
\textbf{Observation} 
The plots of test RMSE delineate the faster training convergence of $g$ over $f$. The hyperbolic expansive convolution operation successfully exploits the underlying intricate hyperbolic structures in the data points. Even in the Euclidean space, the RMSE for $g$ converges faster than the same for $f$, signaling the capacity to exploit the hyperbolic patterns in the data.


\section{Related Works}

\subsection*{Hyperbolic Image Embedding and NLP Tasks}
Developing a Hyperbolic Neural network for computer vision tasks has been mainly focused on combining Euclidean Encoders and Hyperbolic Embedding. These architectures were demonstrated to be effective in performing various vision tasks, for example, recognition \cite{khr},\cite{guo}, generation \cite{nagan}, and image segmentation \cite{atigh}. While Hyperbolic Embedding has also been tremendously successful in performing various tasks related to Natural Language Processing \cite{nickel1},\cite{nickel2}. These ideas were mainly motivated by the expressive power of the hyperbolic spaces to represent graph or tree-like hierarchies in shallow dimensions with very low distortions. However deploying Riemannian Optimization algorithms to train this architecture is difficult due to the inability to extend them for visual data since NLP tasks lack the availability of discrete data \cite{sala1},\cite{sarkar}. 

\subsection*{Fully Connected Hyperbolic Neural Network} 
In 2018 \cite{gan}, and in 2020 \cite{shim} independently developed the structure of Hyperbolic Neural Networks on Poincar\'{e} Disc by utilizing the gyrovector space structure. They defined the generalized notions of different layers like fully connected, convolutional, or attention layers. \cite{fan}, \cite{qu} tried to develop variations of HNN models like fully Hyperbolic GAN on Lorentz Model space, \cite{van} proposed a fully hyperbolic CNN architecture on Poincar\'{e} Disc model. Very recently, \cite{bde} presented a fully convolutional neural network on the Lorentz Model to perform complex computer vision tasks, where they generalized fundamental components of CNNs and proposed novel formulations of convolutional layer, batch normalization, and Multinomial Logistic Regression (MLR) classifier. Moreover, hyperbolic graph neural networks can also accomplish recommending tasks. There are numerous recommender systems such as graph neural collaborative filtering \cite{sun}, \cite{yang}, social network enhanced network system \cite{wang2}, knowledge graph enhanced recommender system \cite{chen2}, and session-based recommender system \cite{guo}, \cite{li2}. 

\subsection*{Batch Normalization in Hyperbolic Neural Networks}
Batch Normalization \cite{batch} restricts the internal departure of neuron outputs by normalizing the outputs produced by the activations at each layer. This adds stability to the training procedure and speeds up the training phase. Several attempts have been made to transcend the normalization of conventional neural networks in the hyperbolic setup. The general framework of Riemannian Batch Normalization \cite{lou}, however, suffers from slower computation and iterative update of the Frech\'{e}t centroid, which does not arise from Gyrovector Group properties. Additionally, \cite{bde} proposed an efficient batch normalization algorithm based on the Lorentz model, utilizing the Lorentz centroid and a mathematical re-scaling operation. 

\subsection*{Numerical Stability of Hyperbolic Neural Networks} Training of Hyperbolic Neural Networks developed on Lorentz Model can lead to instability and floating point error due to rounding since the volume of the Lorentz model grows exponentially with respect to radius. Sometimes, people work with these floating point representations in 64-bit precision with higher memory cost. \cite{numeric1},\cite{numeric2},\cite{numeric3} proposed some versions of feature clipping and Euclidean reparameterization to mitigate these issues. However, they largely overlooked some critical aspects, such as defining a fully hyperbolic convolutional layer or classifiers like MLR, which are essential for various computer vision tasks. In this paper, we fully address this gap by developing a novel architecture from the ground up, along with the theory of its universal consistency.


\section{Preliminaries}
\label{sec:2}

This section discusses the preliminaries of Riemannian Manifolds and Hyperbolic Geometry which would underpin the introduction of our proposed framework. 

\vspace{5pt}
\noindent
\textbf{Riemannian manifold, Tangent space, and Geodesics}
Let us assume an \textit{$n$-dimensional Manifold} $\mathcal{M}$ is a topological space that locally resembles $\rr^n$ \cite{tu}. For each point $x\in\mathcal{M}$, we can define the \textit{Tangent Space} $T_x(\mm)$ as the first order linear approximation of $\mm$ at $x$. We call $\mm$ as a \textit{Reimannian Manifold} if there is a collection of metrics $g:=\{g_x:T_x(\mm)\times T_x(\mm)\to\rr , x\in\mm\}$ at every point of $\mm$ \cite{carmo}.  This metric induces a distance function between two points $p,q\in\mm$ joined by a piecewise smooth curve $\gamma:[a,b]\to\mm$ with $\gamma(a)=p, \gamma(b)=q$ and the distance between $p$ and $q$ is defined as $L(\gamma):=\int_{a}^bg_{\gamma(t)}(\gamma^\prime(t),\gamma^\prime(t))^{1/2}dt$. The notion of \textit{Geodesic} between two such points is meant to be that curve $\gamma$ for which $L(\gamma)$ attains the minimum and that $L(\gamma)$ is referred as the \textit{Geodesic Distance} between $p$ and $q$ in that case. Given such a Riemannian Manifold $\mm$ and two linearly independent vectors $u$ and $v$ at $T_x(\mm)$, we define the sectional curvature at $x$ as $k_x(u,v):=\frac{g_x(R(u,v)v,u)}{g_x(u,u)g_x(v,v)-g_x(u,v)^2}$, where $R$ being the Riemannian curvature tensor defined as $R(u,v)w:=\nabla_u\nabla_v w-\nabla_v\nabla_u w-\nabla_{(\nabla_uv-\nabla_vu)}w$ [$\nabla_uv$ is the directional derivative of $v$ in the direction of $u$, which is also known as the \textit{Rimannian Connection} on $\mm$. ] 

We use these notions to define an $n$-dimensional model \textit{Hyperbolic Space} as the connected and complete Riemannian Manifold with a constant negative sectional curvature.  There are various models in use for Hyperbolic Spaces, such as Poincar\'{e} Disc Model, Poincar\'{e} half Space Model, Klein-Beltrami Model, Hyperboloid Model, etc, but the celebrated \textit{Killing-Hopf Theorem} \cite{lang} asserts that for a particular curvature and dimension, all the model hyperbolic spaces are isometric. This allows us to develop our architecture uniquely (without worrying much about performance variations) over a particular model space, where we choose to work with the Poincar\'{e} Disc model for our convenience. Here, we have briefly mentioned the critical algebraic operations on this model required for our purpose.  

\vspace{5pt}
\noindent
\textbf{Poincar\'{e} Disc Model}
For a particular curvature $k(<0) [c=-k]$, an $n-$ dimensional Poincar\'{e} Disc model contains all of its points inside the ball of radius $1/\sqrt{c}$ embedded in $\rr^n$ \cite{lee}. The geodesics in this model are circular arcs perpendicular to the spherical surface of radius $1/\sqrt{c}$. The geodesic distance between two points $p$ and $q$ (where $\|p\|,\|q\|<1/\sqrt{c}$) is defined as
\begin{align*}
    d(p,q):=2\sinh^{-1}\left(\sqrt{2\frac{\|p-q\|^2}{c(\frac{1}{c}-\|p\|^2)(\frac{1}{c}-\|q\|^2)}}\right),
\end{align*}

From now on, we will denote $\dd^n$ as the $n-$ dimensional Poincar\'{e} Disc with curvature $-c$.  

\vspace{5pt}
\noindent
\textbf{Gyrovector Space}
The concept of Gyrovector Space, introduced by Abraham A. Ungar [see \cite{ungar}], serves as a framework for studying vector space structures within Hyperbolic Space. This abstraction allows for defining special addition and scalar multiplications based on weakly associative gyrogroups. For a detailed geometric formalism of these operations, Vermeer's work \cite{vermeer} provides an in-depth exploration.

In this context, we will briefly discuss M$\ddot{o}$bius Gyrovector Addition and Mobius Scalar Multiplication on the Poincar'{e} Disc. Due to isometric transformations between hyperbolic spaces of different dimensions, the same additive and multiplicative structures can be obtained for other model hyperbolic spaces (refer \cite{rs}). Utilizing M$\ddot{o}$bius addition and multiplication is essential when evaluating intrinsic metrics like the Davies-Bouldin Score or Calinski-Harabasz Index to assess the performance of our proposed algorithm.

\begin{enumerate}
    \item \textbf{M$\ddot{o}$bius Addition:} For two points $u$ and $v$ in the Poincar\'{e} Disc, the M$\ddot{o}$bius addition is defined as:
    $
        u\oplus_c v:=
        \frac{(1+2c<u,v>+c\|v\|^2)u+(1-c\|u\|^2)v}{1+2c<u,v>+c^2\|u\|^2\|v\|^2},$
     where $c$ is the negative of the curvature for hyperbolic spaces and without loss of generality, we assume $c=1$.   

    \item \textbf{M$\ddot{o}$bius Scalar Multiplication:} For $r\in\rr$, $c>0$ and $u$ in the Poincar\'{e} Disc, the scalar multiplication is defined as:
    $
        r\otimes_c u:= \frac{1}{\sqrt{c}} \tanh\left(r \tanh^{-1}(\sqrt{c}\|u\|)\right)\frac{u}{\|u\|}
    $
    This addition and scalar multiplication satisfy the axioms pertaining to the Gyrovector Group [see \cite{ungar}].
\end{enumerate}

\vspace{5pt}
\noindent
\textbf{Exponential \& Logarithmic Maps}
For any $x\in\dd^n$, the $\exp_x^c:T_x(\dd^n)\subseteq \rr^n\to\dd^n$ translates a point in the tangent space of the Poincare Disc and projects it on the Poincar\'{e} Disc along the unit speed geodesic starting from $x\in\dd^n$ in the direction $v\in T_x(\dd^n)$. The Logarithmic map does precisely the opposite, i.e., $\log_x^c:\dd^n\to T_x(\dd^n)\subseteq \rr^n$, projecting a point from the Poincar\'{e} Disc back to the tangent space at $x\in\dd^n$ along the reverse of the geodesic traced by the Exponential Map. Their formulations are explicitly given as follows:
\begin{align*}
    \exp_x^c(v):=x\oplus_c\left(\tanh\left(\sqrt{c}\frac{\lambda_x^c\|v\|}{2}\right)\frac{v}{\sqrt{c}v}\right)
\end{align*}
and 
\begin{align*}
    \log_x^c(z):=\frac{2}{\sqrt{c}\lambda_x^c}\tanh^{-1}\left(\sqrt{c}\|-x\oplus_c y\|\right)\frac{-x\oplus_c y}{\|-x\oplus_c y\|},
\end{align*}
for $y\neq x$ and $v\neq 0$ and the Poincar\'{e} conformal factor $\lambda_x^c:=\frac{2}{(1-c\|x\|^2)}$.

Having all the necessary mathematical preliminaries, we are going to design our model architecture.

\section{Proposed Method}
\label{sec:3}
In this section, we will unravel the design strategy of expansive Hyperbolic Deep Convolutional Neural Networks (eHDCNN). Let us first define the hyperbolic convolution operation on Poincar\'{e} Disc. Assume two functions $f$ and $g$ from $\rr^n\to\rr$, we define the convolution between $f$ and $g$ as: 
\begin{align*}
    f\star g(x):=\int_{\rr^n}f(z)g(x-z)dz.
\end{align*}

Analogously, we define hyperbolic convolution using logarithmic and exponential maps.
\begin{definition}{\bf Hyperbolic Convolution (Continuous Version)}
    For $x\in\dd^n$, we define the convolutions of $f,g:\rr^n\to\rr$ as
    \begin{equation}
        f\star g(x):=\exp_0^c\left[\int_{\dd^n}f(\log_0^c(z))g(\log_0^c(-z\oplus_c x))\lambda(z)\right],
    \end{equation}
    where $\lambda(z):=\frac{dz}{1-c\|z\|^2}$.
\end{definition}
\begin{remark}
    Note that for two real-valued functions $f,g$, their hyperbolic convolution is a map $h:\dd^n\to\dd^1$. We want to keep the range of the output function of the convolution in $\dd$, since in the deep convolutional setup we will again convolute the output with some other filters. 
\end{remark}

\begin{definition}{\bf Hyperbolic Expansive and Contractive Convolution (Discrete Version) }\label{def:4.2}
Let $\boldsymbol{w}:=\{w_j\}_{j=-\infty}^\infty$ be an infinite dimensional vector whose elements are in $\rr$ with finitely many non-zero entries in $\boldsymbol{w}$. Explicitly we assume $w_j\neq 0$ for $0\leq i\leq s$. Among two widely used types of 1-D convolutions in $\rr^n$, we talk about only the expansive and contractive type convolutions. Let $\boldsymbol{v}=\{v_1,...,v_n\}\in \dd^n$. We define the Hyperbolic Expansive Convolution $(\ast_h)$ and the Hyperbolic Contractive Convolution $(\star_h)$ in the following way:\\

Let $\boldsymbol{v^\prime}:=\log_0^c(\boldsymbol{v})=(v_1^\prime,v_2^\prime,...,v_n^\prime)\in \tn_0^c(\dd^n)\subseteq \tn_0^c(\rr^n)$, i.e. $\boldsymbol{v^\prime}$ is an element of the tangent bundle at $0$ of $\dd^n$. 
\begin{figure*}[t]
    \centering
    \includegraphics[width=0.95\textwidth]{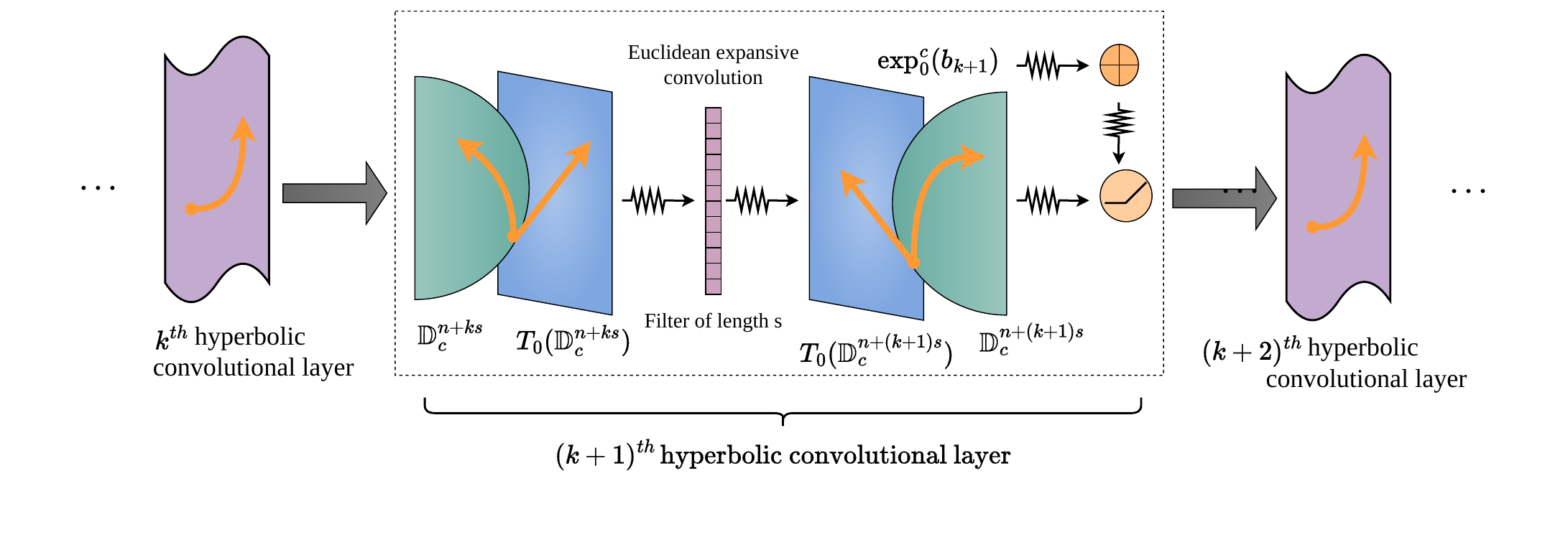}
    \caption{The complete workflow of expansive hyperbolic $1$-D convolutional layer on Poincar\^{e} disc is presented. (best view in digital format) }
    \label{fig:model_overview}
\end{figure*}
\begin{enumerate}
    \item \textbf{Hyperbolic Expansive Convolution:} $(\boldsymbol{w}\ast \boldsymbol{v^\prime})=\sum_{l=1}^nw_{j-l}v_l^\prime$ for $j=1,2,...,n+s$. Therefore  $(\boldsymbol{w}\ast \boldsymbol{v^\prime})\in \rr^{n+s}$. We apply the $\exp$ map to put it back in $\dd^{n+s}$. Finally, we define
    \begin{equation}
        \boldsymbol{w}\ast_h\boldsymbol{v}:=\exp_0^c(\boldsymbol{w}\ast\log_0^c(\boldsymbol{v})).
    \end{equation}
    \item \textbf{Hyperbolic Contractive Convolution:} The usual contractive convolution for $\boldsymbol{w}$ and $\boldsymbol{v^\prime}$ is defined as, $\boldsymbol{w}\star\boldsymbol{v^\prime}=\sum_{l=j-s}^jw_{j-l}v_l, j=s+1,...,n$. We define $\star_h$ between $\boldsymbol{w}$ and $\boldsymbol{v}$ as 
    \begin{equation}\label{eqn:0.10}
        \boldsymbol{w}\star_h\boldsymbol{v}:=\exp_0^c(\boldsymbol{w}\star\log_0^c(\boldsymbol{v})),
    \end{equation}
    which lies in $\dd^{n-s}$.
\end{enumerate}    
\end{definition}

\begin{remark}
    Note that for $c=0$, we will retrieve the usual sparse Toeplitz operators of dimensions $n\times (n+s)$ and $n\times (n-s)$ from those two cases. 
\end{remark}

Having discussed all the necessary terminologies, we are finally set to define the complete architecture of the Hyperbolic Deep Convolutional Neural Network (HDCNN).

\begin{definition}{\bf Hyperbolic Deep Convolutional Neural Network (HDCNN)}
    Let $L\in\mathbb{N}$ be the number of hidden layers in the network. For a given set of filters $\{\boldsymbol{w_k}\}_{k=1}^L$ and set of compatible bias vectors $\{\boldsymbol{b_k}\}_{k=1}^L$ and a vector $\boldsymbol{a_L}=\{a_1,a_2,..., a_{n_L}\}$ [Note that these vectors all lie in Euclidean Spaces of Appropriate Dimensions]. Let $\sigma(t):=\max\{0,t\}$ be the ReLU, acting component-wise for the multidimensional operation. We also assume the dimension of the output layer is $n_L$ and $\boldsymbol{h}_L(x):=\{h^1(x),h^2(x),...,h^{n_L}(x)\}\in\dd^{n_L}$. The HDCNN is defined as:    
    \begin{equation}
        \boldsymbol{h}_k(x)=\sigma(\boldsymbol{w}_{k}\circ \boldsymbol{h}_{k-1}(x)\oplus_c \exp_0^c(\boldsymbol{b}_k)),
    \end{equation}
    where $\boldsymbol{h}_k(x)$ is the output from the $k-$th hidden layer for $k\in\{1,2,...,L-1\}$ and $h_k(x)\in\dd^{n+ks}$, where $\circ$ can be either $\ast_h$ or $\star_h$ as defined in Definition \ref{def:4.2}, $\boldsymbol{h}_0(x)=x$ and the final output is given as:
    \begin{equation}\label{eqn:4.4}
        h_L(x)= \exp_0^c\left[\boldsymbol{a}_L\cdot \log_0^c(\boldsymbol{h}_L(x))\right].
    \end{equation}
    
\end{definition}

The transformation of a vector, lying as a geodesic on Poincar\'{e} Disc, is shown in Figure \ref{fig:model_overview} through hyperbolic convolution performed in an intermediate layer. This transformation projects the vector as another geodesic in a higher dimensional Poincar\'{e} Disc to the subsequent layer. 
\begin{remark}
    If we restrict our focus only to the Expansive case (eHDCNN), note that the dimension of the input to each hidden layer is getting bigger by $s$ units every time. More explicitly, if we have started with $x\in\dd^n$, and $\boldsymbol{w}_1$ is the first filter of length $s$, then $\boldsymbol{h}_1(x)\in\dd^{n+s}$, which is the input dimension of the second hidden layer. Iteratively, the input dimension of the $k-$th hidden layer is as same as the dimension of $\boldsymbol{h}_{k-1}$, which lies in $\dd^{n+(k-1)s}$. Finally, when we reach the output layer, the output dimension will be $n+Ls$, i.e., $n_L=n+Ls$. To make the M$\ddot{o}$ bius addition and the M$\ddot{o}$bius multiplications compatible, we need to have $\boldsymbol{a}_L\in\rr^{n+Ls}$. Also note that for $c=0$, this architecture is reduced to the HDCNN architecture described in \cite{lin}. 
\end{remark}

\section{Theoretical Analyses}
\label{sec:4}

We will now provide the proof for universal consistency following the framework established in \cite{lin}. While we will appropriately generalize the results to the hyperbolic setting, it is first necessary to define some statistical terminologies to comprehend the mechanism of eHDCNN. 

We consider a dataset $\mathcal{D} = \{z_i\}_{i=1}^m = \{x_i, y_i\}_{i=1}^m$, where the samples are assumed to be independent and identically distributed according to a Borel probability measure $\rho$ on the space $\mathcal{Z} = \mathcal{X} \times \mathcal{Y}$. Here, $x_i \in \mathcal{X} \subseteq \dd^n$ and $y_i \in \mathcal{Y} \subseteq \dd^1$. We assume $\mathcal{X}$ is a compact set for this discussion. The goal is to learn a function $f_{\mathcal{D}}: \mathcal{X} \to \mathbb{R}^1$ that minimizes the following \textit{Hyperbolic Generalization Error (HGE)}:

\begin{align*}
    \ec (f)&:=\int_{\mathcal{Z}}(f(x)-\log_0^c(y))^2d\rho.    
\end{align*}
\begin{remark}
    The reason behind taking the $\log$ of $y\in\mathcal{Y}$ is that, the logarithm function will project back $y\in\mathcal{Y}$ to $\mathbb{T}_0(\dd^1)\subseteq \rr^1$. Hence, taking the difference between two real numbers will make sense. Also, if $c\to 0$, we will return the usual generalization error on the Euclidean Spaces. 
\end{remark}

\begin{lemma}
\label{lem8}
    The \textit{Hyperbolic Regression Function (HRF)} $f_\rho(x):=\int_{\mathcal{Y}}\log_0^c(y)d\rho(y|x)$, defined by the means of conditional distribution $\rho(\cdot|x)$ of $\rho$ at $x\in\mathcal{X}$ minimizes the HGE.
\end{lemma}

The next lemma will deduce what we aim to minimize. 
\begin{lemma}\label{lem9}
    For any $f:\dd^n\to\rr^1$, we have
    \begin{align*}
        \ec(f)-\ec(f_\rho)=\|f-f\rho\|_{L^2_{\rho_\xl}},
    \end{align*}
    where $\rho_\xl(x):=\int_\yl \rho(x,y)d\yl(y)$, for each $x\in\xl$, the marginal distribution of $\rho$ on $\xl$.
\end{lemma}

The estimator that minimizes the hyperbolic generalization error is that estimator which minimizes the empirical error over the class of all functions expressed by our eHDCNN architecture. Hence the corresponding estimator or the \textit{Empirical Risk Minimizer (ERM)} is defined as:
\begin{align*}
    f_{\mathcal{D},L,s}:=\arg\min_{f\in\mathcal{H}_{L,s}}\epsilon_\mathcal{D}(f),
\end{align*}
where
\begin{align*}
    \ec_\mathcal{D}(f):=\frac{1}{m}\sum_{i=1}^m\left(f(x_i)-\log_0^c(y_i)\right)^2
\end{align*}
denotes the empirical risk (HERM) associated with the function $f$ and for the filters $\boldsymbol{w_k}$ for $k\in\{1,2,...,L\}$ of length $s_k=d+ks$ and
\begin{align*}
    \mathcal{H}_{L,s}:=\{h_L(x), \boldsymbol{w}_k,\boldsymbol{b}_k\in\rr^{d+ks}, k=1,2,...,L\}
\end{align*}
is the set of all hyperbolic outputs produced by the eHDCNN defined by \ref{eqn:4.4}. 

Now, to verify the consistency, we need to show that when the sample size $m\to\infty$, the sequence of estimators converges to the real estimate. This is formally defined as follows:
A sequence of estimators for a parameter is said to be strongly universally consistent if it converges almost surely to the true value of the parameter. In the case of a regression problem, we say that a sequence of empirical error estimators, built through empirical risk minimizers, is strongly universally consistent if they approach the generalization error over the class of outputs belonging to the Hilbert Space of square-integrable functions with respect to the distribution measure of the output conditioned on the input variable. Therefore in the hyperbolic setup, we define it in the following way: 
\begin{definition}
     A sequence of \textit{Hyperbolic Regression Estimators (HRE)} \(\{f_m\}_{m=1}^\infty\) built through ERM is said to be strongly universally consistent if it satisfies the condition:
    \begin{align*}
        \lim_{m \to \infty} \mathcal{E}(f_m) - \mathcal{E}(f_\rho) = 0
    \end{align*}
     almost surely, for every Borel probability distribution $\lambda$ such that $\log_0^c(\mathcal{Y})\in L^2(\lambda_{(\mathcal{Y}|x)})$.     
\end{definition}

The main result we will be going to prove here will be the following Theorem, which will prove the strong universal consistency of eHDCNN when the Hyperbolic Empirical Risk is minimized. The following Theorem considers a sequence of eHDCNNs as the universal approximators of continuous functions, where the depth of the network has been taken as a sequence depending upon the sample size of our dataset.

\begin{theorem}\label{thm:5.1}
In light of the above discussion, suppose $L=L_m\to\infty$, $M=M_m\to \frac{1}{\sqrt{c}}$ (therefore, $M_m\times \\
\left(\frac{1}{M_m\sqrt{c}}\tanh^{-1}(M_m\sqrt{c})\right)\to\infty$), $m^{-\theta}M_m^2\left[1+\frac{1}{M_m\sqrt{c}}\tanh^{-1}(M_m\sqrt{c})\right]^2\to 0$ [constrained truncation on the power of sample size] and
    \begin{align}\label{eqn 0.11}
        &\frac{\left(\frac{1}{\sqrt{c}}\tanh^{-1}(M_m\sqrt{c})\right)^4 L_m^2(L_m+d)\log(L_m)}{m^{1-2\theta}}\times \log\left(\left(\frac{1}{\sqrt{c}}\tanh^{-1}(M_m\sqrt{c})\right)m\right)\to 0,
    \end{align}
    holds for $\theta\in(0,1/2)$ and for input filter length as $2\leq s\leq d$.  Then $\pi_{M_m}f_{D,L_m,s}$ is strongly universally consistent, where $\pi_M(l):=\min\{M,|l|\}\cdot sign(l)$ is the well known truncation operator. 
\end{theorem}

\begin{remark}
     If we put $\lim c\to 0$ in Theorem \ref{thm:5.1}, we get back Theorem 1 in \cite{lin}. Therefore, Theorem \ref{thm:5.1} is a more generalized version, which is reduced to its Euclidean version for curvature $0$. 
\end{remark}

\begin{remark}
    When we intend to perform the convergence analysis of a series in mathematical analysis, we first consider the partial sum of the series up to a certain term (let's say up to the $k-$th term) and then try to observe the behavior of the series by letting $k\to\infty$. This idea generates the involvement of the truncation operator in Theorem \ref{thm:5.1}. Note that instead of taking $M_m\to\infty$ [which is used in \cite{lin}], we have made $M_m\to\frac{1}{\sqrt{c}}$ (letting our samples lie close to the boundary of the Poincar\'{e} Disc, whose radius is $\frac{1}{\sqrt{c}}$). As $M_m\to\frac{1}{\sqrt{c}}$, $\tanh^{-1}(M_m\sqrt{c})\to\infty$, so does $M_m\left(\frac{1}{M_m\sqrt{c}}\tanh^{-1}(M_m\sqrt{c})\right)$.  It will ease our work for giving an upper bound on the covering number of $\mathcal{H}_{L,s}$ in terms of the truncation limit. Our adoption of the truncation operator is motivated by the widespread application of this operator in proving the universal consistency of various learning algorithms \cite{walk}, \cite{lin}.\\

    Apart from the truncation operator in Theorem \ref{thm:5.1}, several constraints are involved which are crucial to guarantee universal consistency. The constraint on depth $L_m\to\infty$ appears naturally as it is necessary for the universal approximation used in Lemma \ref{lem:19.1}. The growth of the truncation limit concerning sample size $m$ is given by $m^{-\theta}M_m^2\left[1+\frac{1}{M_m\sqrt{c}}\tanh^{-1}(M_m\sqrt{c})\right]^2\to 0$ instead of $M_m^2m^{-\theta}\to 0$ [given in \cite{lin}] to incorporate the growth restriction of sample error in term of two increasing univariate functions $h_1(M_m)h_2(m^{-1})$, where $h_1(x)=x^2\left[1+\frac{1}{x\sqrt{c}}\tanh^{-1}(x\sqrt{c})\right]^2$ and $h_2(x)=x^\theta, \theta>0$. Finally, the constraint in equation \ref{eqn 0.11} will ensure the absolute difference between the generalization error and empirical error goes to $0$, which will be used to prove Lemma \ref{lem:18.2}. 
    
\end{remark}

\begin{remark}
    Theorem \ref{thm:5.1} only demonstrates the universal consistency of the eHDCNN architecture for one-dimensional convolution. The primary restriction comes from the infeasibility of the convolutional factorization that appeared in \cite{har} [also described in \cite{lin}]. Since the analysis in the hyperbolic set-up also relies on the universal approximation for the conventional eDCNN, the question of universal consistency remains open for two or higher-dimensional eHDCNN structures. 
\end{remark}

We now dive into proving Theorem \ref{thm:5.1}. Our main ingredient will be a version of Concentration Inequality [Theorem 11.4,\cite{walk}] after suitably adjusting the upper bound of the metric entropy concerning pseudo-dimension [Lemma 4, \cite{lin}]. Although our approach is similar to \cite{lin} to some extent, we have been able to derive a stronger version of Lemma 6 in \cite{lin} as presented in the proof of Lemma \ref{lem:18.2} in this paper, showing that the truncated empirical error converges to the truncated generalization error much faster in the case of hyperbolic convolution compared to the traditional Euclidean one. This will be established once we present our experimental results in terms of different curvatures (curvature $0$ denotes the experiment has been done using eDCNN).   

To prove Theorem \ref{thm:5.1} we divide our works into three parts as demarcated in \cite{lin} and will develop the appropriate hyperbolic versions of the corresponding results. We begin with expanding the bounds on the covering number for the class of functions defined in \ref{eqn:4.4}. We first need several terminologies. 

Let $\nu$ be a probability measure on $\xx\in\dd^n$. For a function $f:\xx\to\rr$, we set 
\begin{align*}
    \|f\|_{L^p(\nu)}:=\left(\int_{\xx}|f(x)|^p\nu(x)d\xx(x)\right)^{1/p}.
\end{align*} 
Denote by $L^p(\nu)$ the set of all functions with $\|f\|_{L^p(\nu)}<\infty$. For $\mathcal{A}\subseteq L^p(\nu)$, we denote $\mathcal{N}(\epsilon, \mathcal{A}, \|\cdot\|_{L^p(\nu)})$ the covering number of $\mathcal{A}$ in $L^p(\nu)$, which is the least number of balls of radius $\epsilon$ needed to cover up $\mathcal{A}$ with respect to the $\|\cdot\|_{L^p(\nu)}$ metric. In particular we denote $\mathcal{N}_p(\epsilon, \mathcal{A}, x_1^m):=\mathcal{N}_p(\epsilon, \mathcal{A}, \|\cdot\|_{L^p(\nu_m)})$, where $\nu_m$ is the emperical measure for the dataset $x_1^m:=\{x_1,x_2,...,x_m\}\in\xx^m$. Further we define$\mathcal{M}(\epsilon, \mathcal{A}, \|\cdot\|_{L^p(\nu)})$ to be the $\epsilon-$packing number of $\mathcal{A}$ with respect to the $\|\cdot\|_{L^p(\nu)}$ norm, which is the largest integer $N$ such that given any subset $\{g_1,g_2,...,g_N\}$ of $\mathcal{A}$ satisfies $\|g_i-g_j\|\geq \epsilon$ for all $1\leq i<j\leq N$. 

Next, we will mention lemma 9.2 from \cite{walk}, which expresses a relation involving inequalities among the covering and packing numbers. 
\begin{lemma}\label{lem15}
    Let $\mathcal{G}$ be a class of functions from $\xx\to\rr$ and $\nu$ be a probability measure on $\xx$. For $p\geq 0$ and $\epsilon>0$, we have
    \begin{align*}
        &\mathcal{M}(2\epsilon, \mathcal{G}, \|\cdot\|_{L^p(\nu)})\leq \mathcal{N}(\epsilon, \mathcal{G}, \|\cdot\|_{L^p(\nu)})\leq \mathcal{M}(\epsilon, \mathcal{G}, \|\cdot\|_{L^p(\nu)}).
    \end{align*}
    In particular,
    \begin{align*}
        \mathcal{M}_p(2\epsilon, \mathcal{G}, x_1^m) \leq \mathcal{N}_p(\epsilon, \mathcal{G}, x_1^m) \leq \mathcal{M}_p(\epsilon, \mathcal{G}, x_1^m).
    \end{align*}
\end{lemma}

Next, we have to derive an estimate of the upper bound of the Packing number for the hyperbolic pseudo dimension. 

Since the Lemma $2$, $3$, and $4$ from Capacity Estimates in Appendix A of \cite{lin} are taken from results proved on general metric spaces, we will just state Lemma $4$ from \cite{lin} without proof in the context of hyperbolic space, which we will use later. 

\begin{lemma}\label{lem:16.2}
    For $0<\epsilon\leq M$ and $c^\ast$ being an absolute constant, we have
    \begin{align*}
        &\log_2 \sup_{x^1_m\in\xx^m}\mathcal{N}_1\left(\epsilon, \pi_M\mathcal{H}_{L,s},x_1^m\right)\leq c^\ast L^2(Ls+d)\log(L(s+d))\log\frac{M}{\epsilon}.
    \end{align*}
\end{lemma}

We define the hyperbolic version of the generalization error (HGE) as 
\begin{align*}
    \ec_{\pi_M}(f):=\int_{\mathcal{Z}}\left(f(x)-\log_0^c(y_M)\right)^2d\rho,
\end{align*}
and the \emph{Hyperbolic Empirical Error (HEE) } (truncated) as
\begin{align*}
    \ec_{\pi_M, D}(f):=\frac{1}{m}\sum_{i=1}^m\left(f(x_i)-\log_0^c(y_{i,M})\right)^2,
\end{align*}
where $l_M:=\min\{M,|l|\}\cdot sign(l)$, the well known truncation operator. \\

We now provide a convergence criterion for the HEE estimates to the HGE estimate. We will use a hyperbolic version of the concentration inequality as given in Lemma 5, \cite{lin}. 

A more generalized version of Theorem 11.4 \cite{walk} can be presented as follows:

\begin{lemma}\label{lem:18.1}
    We assume $|y|\leq B$ and $B\geq \frac{1}{\sqrt{c}}$. For a set of functions $\mathcal{F}$ from $f:\xx\to\rr$ satisfying $|f(x)|\leq B$ and for all $m\geq 1$, we have
    \begin{align*}
        & \mathbb{P}[\exists f\in \mathcal{F} : \epsilon(f)-\epsilon(f_\rho)-(\epsilon_D(f)-\epsilon_D(f_\rho))\geq \epsilon(\alpha+\beta+\epsilon(f)-\epsilon(f_\rho))] \\
        &\leq 14 \sup_{x_1^m\in \xx^m} \mathcal{N}_1\left(\frac{\beta\epsilon}{20B},\mathcal{F},x_1^m\right)\exp\left(-\frac{\epsilon^2(1-\epsilon)\alpha m}{214(1+\epsilon)B^4}\right), 
    \end{align*}

    where $\alpha,\beta>0$ and $\epsilon\in(0,1/2)$.     
   
\end{lemma}

Based on Lemma \ref{lem:18.1}, the following Lemma will lay out the convergence criterion of the Truncated HEE estimates.

\begin{lemma}\label{lem:18.2}
    When $m^{-\theta}M_m^2\left[1+\frac{1}{M_m\sqrt{c}}\tanh^{-1}(M_m\sqrt{c})\right]^2 \to 0$ and equation \ref{eqn 0.11} holds for $\theta\in(0,1/2)$, then we have
    \begin{align*}
        \lim_{m\to\infty} \ec_{\pi_{M_m}}(\pi_{M_m}f_{D,L,s})-\ec_{\pi_{M_m},D}(\pi_{M_m}f_{D,L,s})=0
    \end{align*}
    holds almost surely. 
\end{lemma}

We are finally in a position to prove Theorem \ref{thm:5.1}; we will give our final lemma, which will complete the proof for universal consistency. 

\begin{lemma}\label{lem:19.1}
    Let $\Omega\subseteq \dd^d$ be compact and $2\leq s\leq d$. Then for any $f\in\mathcal{C}(\Omega)$, there exist a sequence of filters $\boldsymbol{w}$ and bias vectors $\boldsymbol{b}$ of appropriate dimensions and $f_L^{\boldsymbol{w},\boldsymbol{b}}\in\mathcal{H}_{L,s}$ such that 
    \begin{align*}
        \lim_{L\to\infty}\|f-f_L^{\boldsymbol{w},\boldsymbol{b}}\|_{\mathcal{C}(\Omega)}=0.
    \end{align*}
\end{lemma}

\begin{remark}\label{rem:final}
    We notice from the proof of Lemma \ref{lem:18.2} that the truncated HEE estimates converge much faster to the corresponding HGE than their Euclidean equivalents. This property gives the eHDCNN architecture an edge over the eDCNN for faster training with much fewer training iterations needed. Roughly speaking, since each layer is taking input from a Poincar\'{e} Disc, which in turn expresses the complex representation of the data to the next layer even before the information gets carried out to the next layer directly from the previous layer, the architecture is very quick to learn the internal representation of the data. This will be evident from our simulation results, showing the ascendancy of our architecture over its Euclidean version to achieve lower error rates much faster for certain regression problems.  
\end{remark}

\section{Experiments \& Results}
We will demonstrate the efficacy of eHDCNN by conducting experiments on both synthetic and real-world datasets. All hyperparameters regarding the simulations can be found in the Supplementary document. Our Python-based implementation is available at \href{https://anonymous.4open.science/r/eHDCNN-3C9E/README.md}{https://anonymous.4open.science/r/eHDCNN-3C9E/README.md}.

\begin{figure*}[!ht]
    \centering
    \includegraphics[width=0.95\textwidth]{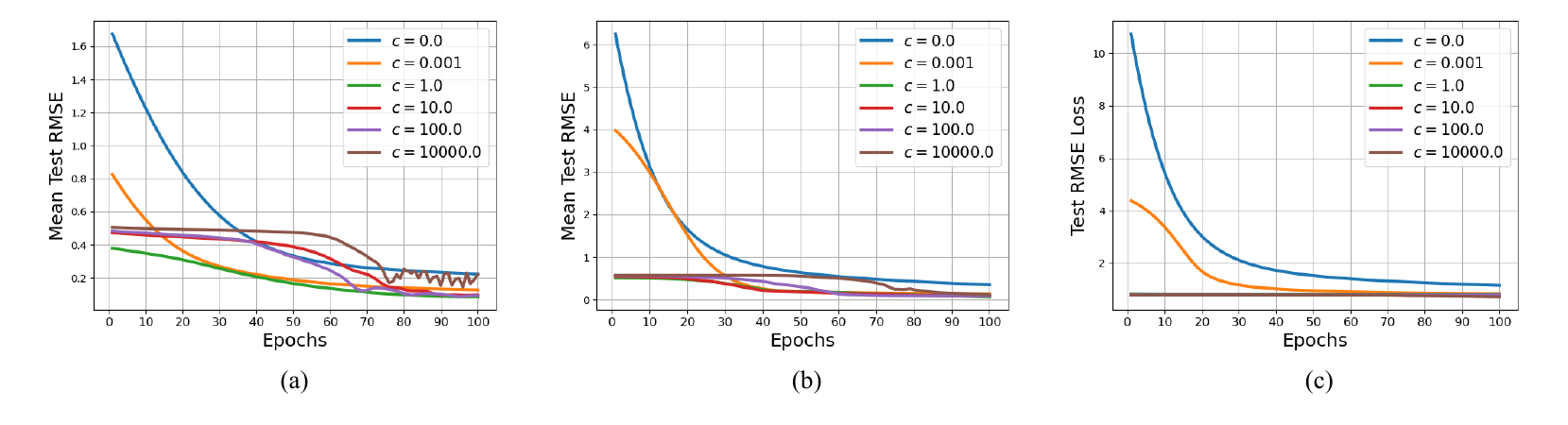}
    \caption{ The performance analysis of eHDCNN with varying space curvatures (a) for $f(x)$ and (b) for $g(x)$, and (c) House price prediction is demonstrated. The Root Mean Square Error (RMSE) decreases faster with increasing curvature, justifying the utility of applying hyperbolic convolution. (best view in digital format)}
    \label{fig:set_1}
\end{figure*}

\begin{figure*}
    \centering
    \includegraphics[width=0.90\textwidth]{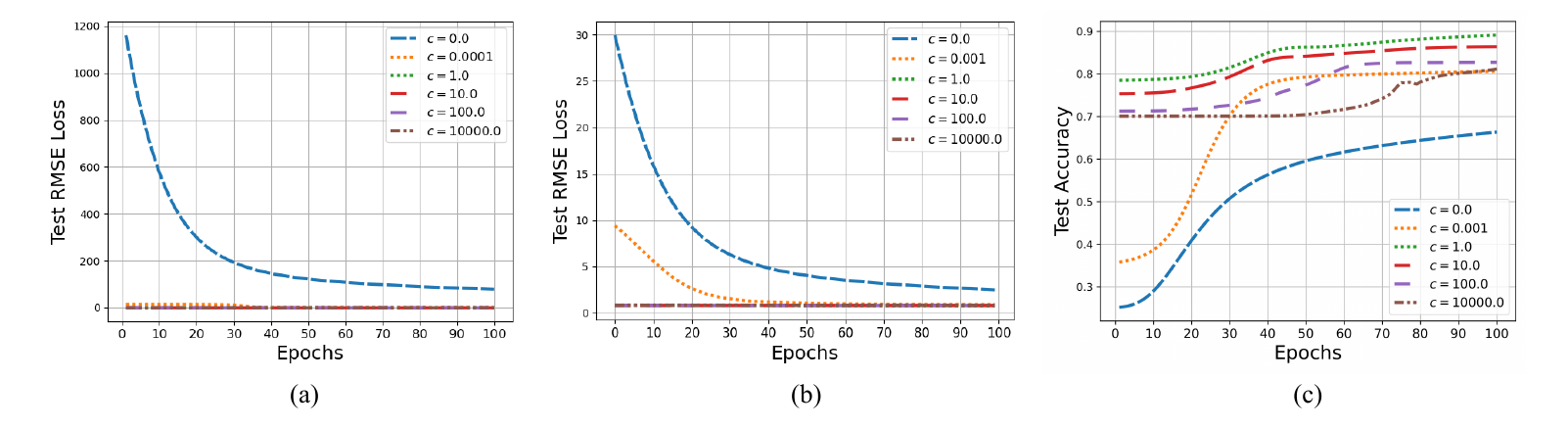}
    \caption{The performance analysis of eHDCNN with varying space curvatures (a) for Superconductivity (b) for Wave Energy, and (c) test accuracy for WISDM are demonstrated. The Root Mean Square Error (RMSE) decreases faster for both (a) and (b) with increasing curvature. On the contrary, test accuracy increases in (c), justifying the utility of employing hyperbolic convolution. (best view in digital format)}
    \label{fig:set_2}
\end{figure*}

\subsection{Synthetic Datasets}
We will construct two regression tasks based on the following functions,
\begin{equation*}
    \begin{split}
        & f(x) = \frac{\sin(\|x\|_2)}{\|x\|_2}, 
        \hspace{1cm} g(x) = \frac{\sqrt{\|x\|_2}}{1+\sqrt{\|x\|_2}}.
    \end{split}
\end{equation*}
We used the regression model $y=h(x)+\epsilon$ (where $h$ can be either $f$ or $g$) to generate the training samples, where $\epsilon\sim \mathcal{N}(0,0.01)$ and $x\sim unif([-1,1]^{10})$. A fixed set of $800/200$ samples for train/test split are used for the experiment, except that the test data are taken without the Gaussian noise. We have used a filter size of length $8$ and the number of layers $4$. We have trained our model over $100$ iterations for each of the $800$ samples and recorded the mean RMSE. We repeat the experiments for six different sets of curvatures. Refer to Figures \ref{fig:set_1}(a) and \ref{fig:set_1}(b) for the detailed illustration. 

The curves are evidence of the faster convergence of test RMSE loss during the entire training process, which validates the Remark \ref{rem:final}. The loss curves are much steeper when the curvatures are more significant than zero compared to the same of the Euclidean counterpart. One point should be noted that the performance of eHDCNN started exacerbating with the higher curvature value. The phenomenon can be attributed to the contraction of Poincar\'{e} Disc with a very high curvature. Thus, the loss curves seem to be overlapping. Yet, the performance is commendable when eHDCNN is trained in the hyperbolic space of low curvature.   

\begin{table*}[!ht]
\centering
\caption{The details of four real-world datasets are presented.} 
\label{tab:dataset_details}
\begin{tabular}{lcccc}
\toprule
Dataset & Superconductivity & Wave Energy Converters & House Price Prediction & WISDM \\
\midrule
No of samples & $288000$ & $21263$ & $545$ & $1073120$ \\
No. of features & $81$ & $81$ & $12$ & $3$\\
No. of classes & - & - & - & $6$ \\
Target task & Regression & Regression & Regression & Classification \\
\bottomrule
\end{tabular}
\end{table*}

\subsection{Real-world Datasets}
We considered four real-world datasets to showcase the effectiveness of eHDCNN. The details of the datasets and the hyperparameters are provided respectively in Table \ref{tab:dataset_details} and \ref{tab:hyper_details}. 

\begin{table*}[!ht]
\centering
\caption{The complete details of hyperparameters for four real-world datasets are presented to reproduce the results. }
\label{tab:hyper_details}
\begin{tabular}{lcccc}
\toprule
Hyperparameters  &  Superconductivity &   Wave Energy Converters  &  House Price Prediction & WISDM \\
\midrule
No of layers & $4$ & $4$ & $4$  & $4$\\
length of input filter & $8$ & $8$ & $8$ & $9$\\
Noise  & No & No & No  & No\\
Learning Rate & $0.01$ & $0.01$ & $0.01$ & $0.01$ \\
Weight decay & $0.0005$ & $0.0005$ & $0.0005$  & $0.0005$\\
Train/test split & $0.80$ & $0.80$ & $0.80$ & $0.70$\\
No of samples & $288000$ & $21263$ & $545$ & $1073120$ \\
Input dimension & $81$ & $81$ & $12$ & $240$\\
Batch Size & $128$ & $128$ & Full & $128$\\
\bottomrule
\end{tabular}
\end{table*}

\begin{figure*}[!ht]
    \centering
    \includegraphics[width=1.01\textwidth]{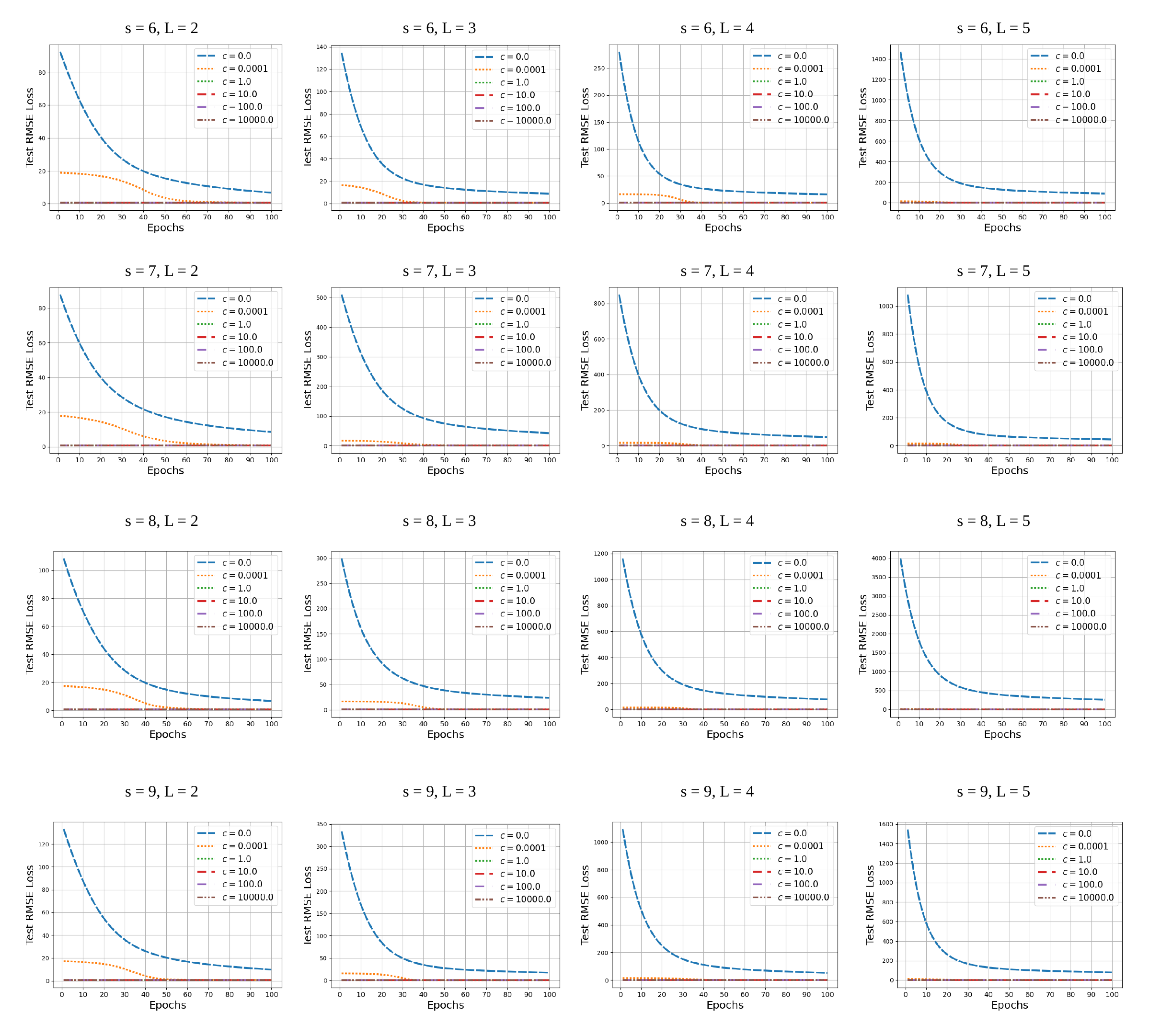}
    \caption{Various experiments were performed on the Superconductivity dataset by varying filter length and number of convolutional layers of the eHDCNN architecture. }
    \label{fig:ablation}
\end{figure*} 

\subsubsection{Regression Task}
We include $3$ real-world regression datasets to demonstrate the performance of eHDCNN over the prevailing DCNN. We deploy the same eHDCNN architecture with $4$ layers, and the length of the input filter is $8$ for all three regression tasks. We split the entire dataset into $80\%$ samples for training and the rest $20\%$ samples for testing. We record the standardized test RMSE over the number of iterations during the training phase. 

\vspace{5pt}
\noindent
\textbf{House Price Prediction}
We consider the widely available house price prediction dataset \cite{house} to solve the regression task. This dataset consists of $545$ samples with $12$ input features such as area, number of bedrooms, furnishing status, air conditioning, etc.  At first, we standardize the entire data after numerically encoding its categorical column. We have trained our model using $4$ layers and with an input filter length of $8$. The test RMSE has been plotted against training iterations for six different curvatures in \ref{fig:set_1}(c), where the curvature $0$ means that the test RMSE has been plotted based on the eDCNN model. 


\vspace{5pt}
\noindent
\textbf{Superconductivity}
As described in \cite{superconductivity}, this dataset contains $21263$ samples, each with $81$ features like mean atomic mass, entropy atomic mass, mean atomic radius, entropy valence etc, along with the output feature as the critical temperature in the $82$nd column. We split the dataset into $80:20$ for our training and testing purposes. We will train our model with a mini-batch of size $128$ in each training iteration. The test RMSE has been plotted against the number of training iterations for six different curvatures in \ref{fig:set_2} (a), where the curvature $0$ indicates that the test RMSE has been taken based on the eDCNN model. 

\vspace{5pt}
\noindent
\textbf{Wave-Energy Converters}
As described in \cite{wave-energy}, this dataset contains $288000$ samples, each with $81$ features. This data set consists of positions and absorbed power outputs of wave energy converters (WECs) in four real wave scenarios from the southern coast of Australia (Sydney, Adelaide, Perth, and Tasmania). We split the dataset into $80:20$ for our training and testing purposes. Similar to the Superconductivity dataset, we will train our model with a mini-batch of size $128$ each time. For the test RMSE plot against the number of training epochs, we refer to \ref{fig:set_2} (b).

\subsubsection{Classification Task}
The only dataset we include for solving classification tasks is WISDM. 

\vspace{5pt}
\noindent
\textbf{WISDM}
We have applied eHDCNN on the WISDM, a well-adopted Human Activity Recognition (HAR) dataset \cite{wisdm}. As it is described in \cite{lin}, this dataset has six types of human activities such as cycling, jogging, sitting, standing, going upstairs and downstairs, with the corresponding accelerations along $x,y$, and $z$ axes at different timestamps and several user id ranging from $1$ to $36$. We have used the user IDs from $1$ to $28$ for training and the rest for testing. We have put $80$ consecutive timestamps for each of the six classes together to make our input dimension $80\times 3=240$. After this conversion, our training dataset has $10172$ samples, and the test dataset has $3242$ number of samples. Our experiment is carried out on a network with $4$ layers with input filter length as $9$. We have trained our model with a mini-batch of size $128$ in each epoch.


\vspace{5pt}
\noindent
\textbf{Discussion}
We run experiments on the House Price Prediction, Superconductivity, Wave-Energy Converters, and WISDM and plots can be seen respectively in Figures \ref{fig:set_1}(c), \ref{fig:set_2}(a), \ref{fig:set_2}(b), and \ref{fig:set_2}(c). Test RMSE loss is the metric for the first three datasets, and test accuracy is the metric for the last one. The plots elucidate that the corresponding metric performs better when the curvature increases than the Euclidean variant. The better performance underscores the efficacy of hyperbolic architecture dominates over its Euclidean counterpart. One common point is that performance further degrades when the value of the curvature lies in a very higher range. It occurs due to the shrinkage of the Poincar\^{e} disc with a very high value of the curvature.


\subsection{Ablation Study}
We conduct an ablation study to study the effect of the filter length and number of hidden layers of the eHDCNN. The experiment is performed on Superconductivity. The filter length and number of layers are chosen respectively from the sets $s=\{6,7,8,9\}$ and $L=\{3,4,5,6\}$. We run experiments for each pair of $(s, L)$, and vary curvatures of the Poincar\^{e} disc. The test RMSE curves are plotted and all results are presented in Figure \ref{fig:ablation}. It can be observed that the test RMSE slowly decreases during the initial epochs of training of the eHDCNN. If we increase the number of layers or the length of the input filter, the respective error rates seem to be more stable and converge faster for the eHDCNN. This emphasizes the stability of our proposed architecture during training and is a clear indication of the fact that it requires a much lesser number of training iterations compared to the conventional eDCNN architecture for convergence.

\section{Conclusion \& Future Works}
In this paper, we have identified the limitations of Euclidean spaces in providing meaningful information for training conventional DCNNs. We demonstrated the superiority of hyperbolic convolutions by treating the output of each layer as elements of the Poincar'{e} Disc, projecting them onto the Tangent Space for expansive convolution, and then mapping them back to a higher-dimensional Poincar\'{e} Disc to capture complex hierarchical structures to the next layer. Our primary contribution is the proof of universal consistency by defining regression and error estimators in the hyperbolic space, drawing an analogy to Euclidean space. This is the first known result to explore the statistical consistency of architectures developed beyond Euclidean spaces. Furthermore, our simulation results validate our theoretical justification, showing why eHDCNN is more adept at capturing complex representations, as noted in Remark \ref{rem:final}. We anticipate that our findings will significantly accelerate the growth of deep learning spanning across the hyperbolic regime.

\appendix

\section{Proofs}

\noindent
\textbf{Lemma \ref{lem8}}
   The \textit{Hyperbolic Regression Function (HRF)} $f_\rho(x):=\int_{\mathcal{Y}}\log_0^c(y)d\rho(y|x)$, defined by the means of conditional distribution $\rho(\cdot|x)$ of $\rho$ at $x\in\mathcal{X}$ minimizes the \textit{Hyperbolic Generalization Error (HGE)}.


\begin{proof}
    The HGE can be written in terms of conditional expectation in the following way:
    \begin{align*}
        \ec(f)&=\int_{\mathcal{Z}}(f(x)-\log_0^c(y))^2d\rho\\
        &=\mathbb{E}_{\mathcal{X},\mathcal{Y}}[f(\mathcal{X})- \log_0^c(\mathcal{Y})]^2
    \end{align*}
    Now for any function $g:\mathcal{X}\to\rr^1$, we write
    \begin{align*}
        \ec(g) =& \ee_{\xl}\left[\ee_{\yl|\xl}\left[ \left(g(\xl)-\ee[\log_0^c(\yl)|\xl]+\ee[\log_0^c(\yl)|\xl]-\log_0^c(\yl)\right)^2|\xl\right] \right]    \\
        = & \ee_{\xl}\left[\ee_{\yl|\xl}\left[\left(g(\xl)-\ee[\log_0^c(\yl)|\xl]\right)^2|\xl\right]\right]+\ee_{\xl}\left[\ee_{\yl|\xl}\left[\left(\ee[\log_0^c(\yl)|\xl]-\log_0^c(\yl)|\xl\right)^2|\xl\right]\right]\\
        & +2\ee_{\xl}\left[\ee_{\yl|\xl}\left[\left(g(\xl)-\ee[\log_0^c(\yl)|\xl]\right)\left(\ee[\log_0^c(\yl)|\xl]-\log_0^c(\yl)\right)|\xl\right]\right].
    \end{align*}
    The cross term in the last expression is $0$, since 
    \begin{align*}
        \ee_{\xl}\left[\ee_{\yl|\xl}\left[\left(\ee[\log_0^c(\yl)|\xl]-\log_0^c(\yl)\right)\right]\right]=0.
    \end{align*}
    Therefore, the expression for HGE is reduced to
    \begin{align*}
        \ec(g) = \ee_{\xl}\left[\ee_{\yl|\xl}\left[\left(g(\xl)-\ee[\log_0^c(\yl)|\xl]\right)^2|\xl\right]\right]+\ee_{\xl}\left[\ee_{\yl|\xl}\left[\left(\ee[\log_0^c(\yl)|\xl]-\log_0^c(\yl)|\xl\right)^2|\xl\right]\right],
    \end{align*}
    which attains minimum when $g(x)=\ee\left[\log_0^c(\yl)|x\right]$ for each $x\in\xl$. Alternately, we write for each $x\in\xl$
    \begin{align*}
        g(x)=\int_{\yl}\log_0^c(y)d\rho(y|x). 
    \end{align*}
\end{proof}

\noindent
\textbf{Lemma \ref{lem9}}
    For any $f:\dd^n\to\rr^1$, we have
    \begin{align*}
        \ec(f)-\ec(f_\rho)=\|f-f\rho\|_{L^2_{\rho_\xl}},
    \end{align*}
    where $\rho_\xl(x):=\int_\yl \rho(x,y)d\yl(y)$, for each $x\in\xl$, the marginal distribution of $\rho$ on $\xl$.
\begin{proof}
Following the proof of Lemma \ref{lem8}, we can write
\begin{align*}
    \ec(f)-\ec(f_\rho) & =\ee_\xl\left[\ee_{\yl|\xl}\left[\left(f(x)-\ee[\log_0^c(\yl)|\xl]\right)^2|\xl\right]\right]  \\
    & = \ee_{\xl,\yl}[\left[\left(f(x)-\ee[\log_0^c(\yl)|\xl]\right)^2|\xl\right]\\
    & = \int_\xl \int_\yl (f(x)-f_\rho(x))\rho(x,y)d\xl(x)d\yl(y)\\
    & = \int_{\xl}(f(x)-f_\rho(x))^2\int_{\yl}\rho(x,y)d\yl(y)d\xl(x)\\
    & = \int_\xl (f(x)-f_\rho(x))^2\rho_\xl(x)d\xl(x)\\
    & = \|f-f_\rho\|_{L^2_{\rho_\xl}}.
\end{align*}    
\end{proof}

\noindent
\textbf{Lemma \ref{lem15}} 
Let $\mathcal{G}$ be a class of functions from $\xx\to\rr$ and $\nu$ be a probability measure on $\xx$. For $p\geq 0$ and $\epsilon>0$, we have
    \begin{align*}
        &\mathcal{M}(2\epsilon, \mathcal{G}, \|\cdot\|_{L^p(\nu)})\leq \mathcal{N}(\epsilon, \mathcal{G}, \|\cdot\|_{L^p(\nu)})\leq \mathcal{M}(\epsilon, \mathcal{G}, \|\cdot\|_{L^p(\nu)}).
    \end{align*}
    In particular,
    \begin{align*}
        \mathcal{M}_p(2\epsilon, \mathcal{G}, x_1^m) \leq \mathcal{N}_p(\epsilon, \mathcal{G}, x_1^m) \leq \mathcal{M}_p(\epsilon, \mathcal{G}, x_1^m).
    \end{align*}
\begin{proof}
    The same proof mentioned in Lemma 9.2 \cite{walk}, can be applied to any general metric space $M$ instead of $\rr^d$. In particular $M$ can be $\xx$. This shows the lemma is unaltered in the case of a compact subset in a hyperbolic space. 
\end{proof}

\noindent
\textbf{Lemma \ref{lem:18.2}}
When $m^{-\theta}M_m^2\left[1+\frac{1}{M_m\sqrt{c}}\tanh^{-1}(M_m\sqrt{c})\right]^2\to 0$ and \ref{eqn 0.11} holds for $\theta\in(0,1/2)$, then we have
    \begin{align*}
        \lim_{m\to\infty} \ec_{\pi_{M_m}}(\pi_{M_m}f_{D,L,s})-\ec_{\pi_{M_m},D}(\pi_{M_m}f_{D,L,s})=0
    \end{align*}
    holds almost surely. 
\begin{proof}
    We have $|\pi_{M}f,{D,L,s}|\leq M$ and $|\log_0^c(y_M)|, |\log_0^c(y_{i,M})|\leq \frac{1}{\sqrt{c}}\tanh^{-1}(M\sqrt{c})$ [The last two inequalities follow from the fact that $\tanh^{-1}$ is increasing on $(-1,1)$]. A little computation will show that
    \begin{align*}
        |\ec_{\pi_M}(\pi_Mf_{D,L,s})|\leq M^2\left[1+\frac{1}{M\sqrt{c}}\tanh^{-1}(M\sqrt{c})\right]^2.
    \end{align*}
    This leads us to derive that
    \begin{align*}
        |\ec_{\pi_M}(\pi_Mf_{D,L,s})-\ec_{\pi_M,D}(\pi_Mf_{D,L,s})|\leq 2 M^2\left[1+\frac{1}{M\sqrt{c}}\tanh^{-1}(M\sqrt{c})\right]^2.
    \end{align*}
    Putting $\alpha=\beta=1,$ in Lemma \ref{lem:18.1} and $\epsilon=m^{-\theta}$ we get that
    \begin{align*}
        \ec_{\pi_M}(\pi_Mf_{D,L,s})-\ec_{\pi_M}(f_\rho) - \left(\ec_{\pi_{M,D}}(\pi_Mf_{D,L,s})-\ec_{\pi_{M,D}}(f_\rho)\right) \leq 2m^{-\theta}\left[1+M^2\left(1+\frac{1}{M\sqrt{c}}\tanh^{-1}(M\sqrt{c})\right)\right]^2
    \end{align*}
    holds with probability at least 
    \begin{align*}
        1-14\sup_{x_1^m\in \xx^m} \mathcal{N}_1\left(\frac{1}{20\frac{1}{\sqrt{c}}\tanh^{-1}(M\sqrt{c})m^\theta},\mathcal{F},x_1^m\right)\exp\left(-\frac{m^{1-2\theta}}{428(1+\epsilon)\left(\frac{1}{\sqrt{c}}\tanh^{-1}(M\sqrt{c})\right)^4}\right)
    \end{align*}
    From lemma Lemma \ref{lem:16.2} we write
    \begin{align*}
        &\sup_{x_1^m\in \xx^m} \mathcal{N}_1\left(\frac{1}{20\frac{1}{\sqrt{c}}\tanh^{-1}(M\sqrt{c})m^\theta},\mathcal{F},x_1^m\right)\exp\left(-\frac{m^{1-2\theta}}{428(1+\epsilon)\left(\frac{1}{\sqrt{c}}\tanh^{-1}(M\sqrt{c})\right)^4}\right)\\
        &\leq \exp\left(c^*\log\left(20\left(\frac{1}{\sqrt{c}}\tanh^{-1}(M\sqrt{c})\right)^2m^\theta\right)L_m^2(d+sL_m)\log(L_m(s+d))-\frac{m^{1-2\theta}}{428\left(\frac{1}{\sqrt{c}}\tanh^{-1}(M\sqrt{c})\right)^4}\right)
    \end{align*}
    The conditions of Theorem \ref{thm:5.1} indicates that the RHS of the last inequality goes to $0$ as $m\to\infty$. Combining together with the strong law of large numbers we get
    \begin{align*}
        \ec_{\pi_{M_m}}\left(\pi_{M_m}f_{D,L_m,s}\right)-\ec_{\pi_{M_m},D}\left(\pi_{M_m}f_{D,L_m,s}\right)\leq 2m^{-\theta}\left[1+M  \left(1+\frac{1}{M\sqrt{c}}\tanh^{-1}(M\sqrt{c})\right)\right]^2 \leq 8M^2m^{-\theta} \to 0
    \end{align*}
    as $m\to\infty$ holds almost surely, completing the proof of Lemma \ref{lem:18.2}.
\end{proof}





\noindent
\textbf{Lemma \ref{lem:19.1}}
Let $\Omega\subseteq \dd^d$ be compact and $2\leq s\leq d$. Then for any $f\in\mathcal{C}(\Omega)$, there exist a sequence of filters $\boldsymbol{w}$ and bias vectors $\boldsymbol{b}$ of appropriate dimensions and $f_L^{\boldsymbol{w},\boldsymbol{b}}\in\mathcal{H}_{L,s}$ such that 
    \begin{align*}
        \lim_{L\to\infty}\|f-f_L^{\boldsymbol{w},\boldsymbol{b}}\|_{\mathcal{C}(\Omega)}=0.
    \end{align*}
\begin{proof}
    Define $g(y):=f(\exp_0^c(y))$ for $y\in\log_0^c(\Omega)$. Then, by Theorem 1 \cite{har}, we know that there exists $g_L^{\boldsymbol{w},\boldsymbol{b}}$ [where $g_L^{\boldsymbol{w},\boldsymbol{b}}$ lies in the free parameter space of the DCNN], such that 
    \begin{align*}
        \lim_{L\to\infty}\|g-g_L^{\boldsymbol{w},\boldsymbol{b}}\|_{\mathcal{C}(\log_0^c(\Omega))}=0.
    \end{align*}
    We now define $f_L^{\boldsymbol{w},\boldsymbol{b}}(x):=g_L^{\boldsymbol{w},\boldsymbol{b}}(\log_0^c(x))$ for $x\in\dd^d$. Now it is easy to verify that
    \begin{align*}
        \lim_{L\to\infty}\|f-f_L^{\boldsymbol{w},\boldsymbol{b}}\|_{\mathcal{C}(\Omega)}=\lim_{L\to\infty}\|g\circ\log_0^c-g_L^{\boldsymbol{w},\boldsymbol{b}}\circ\log_0^c\|_{\mathcal{C}(\log_0^c(\Omega))}=0,
    \end{align*}
    since the $\log_0^c$ [hence its inverse $\exp_0^c$] is global diffemorphism from $\dd^d\to\rr^d$ [from $\rr^d\to\dd^d$]. 
\end{proof}



   
 \noindent
 \textbf{Theorem \ref{thm:5.1}}
 Suppose $L=L_m\to\infty$, $M=M_m\to \frac{1}{\sqrt{c}}$ (therefore, $M_m\times \\
\left(\frac{1}{M_m\sqrt{c}}\tanh^{-1}(M_m\sqrt{c})\right)\to\infty$), $m^{-\theta}M_m^2\left[1+\frac{1}{M_m\sqrt{c}}\tanh^{-1}(M_m\sqrt{c})\right]^2\to 0$ [constrained truncation on the power of sample size] and
    \begin{align}
        &\frac{\left(\frac{1}{\sqrt{c}}\tanh^{-1}(M_m\sqrt{c})\right)^4 L_m^2(L_m+d)\log(L_m)}{m^{1-2\theta}}\times \log\left(\left(\frac{1}{\sqrt{c}}\tanh^{-1}(M_m\sqrt{c})\right)m\right)\to 0,
    \end{align}
    holds for $\theta\in(0,1/2)$ and for input filter length as $2\leq s\leq d$.  Then $\pi_{M_m}f_{D,L_m,s}$ is strongly universally consistent, where $\pi_M(l):=\min\{M,|l|\}\cdot sign(l)$ is the well known truncation operator. 
 
\begin{proof}
    We have $\mathbb{E}[(\log_0^c(y))^2]<\infty$, i.e. $f_\rho\in L^2(\rho_\xx)$. By Lemma \ref{lem:19.1}, we say that there exists a big enough $L_\epsilon$ so that $f_{L_\epsilon}^{w,b}\in\mathcal{H}_{L_\epsilon,s}$ with
    \begin{align*}
        \|f_\rho-f_{L_\epsilon}^{w,b}\|^2_{L^2(\rho_xx)}\leq\left[\limsup_{x\in\xx}\|f_\rho(x)-f_{L_\epsilon}^{w,b}(x)\|\right]^2=\left[\|f_\rho-f_{L_\epsilon}^{w,b}\|_{\mathcal{C}(\xx)}\right]^2\leq\epsilon,
    \end{align*}
    where the second inequality follows from the fact that $\rho_\xx$ being a Borel Probability measure on $\xx$.

    By triangle inequality, we write
    \begin{align*}
        &\ec(\pi_M(f_{D,L,s}))-\ec(f_\rho)\\
        \leq & \epsilon(\pi_M(f_{D,L,s}))-(1+\epsilon)(\pi_M(f_{D,L,s}))\\
        + & (1+\epsilon)\left(\ec_{\pi_M}(\pi_M(f_{D,L,s}))\right)-\ec_{\pi_M,D}(\pi_M(f_{D,L,s}))\\
        + & (1+\epsilon)\left(\ec_{\pi_{M},D}(\pi_M(f_{D,L,s}))-\ec_{\pi_M,D}(f_{D,L,s})\right)\\
        + & (1+\epsilon)(\ec_{\pi_{M},D}(f_{D,L,s}))-(1+\epsilon)^2(\ec_{D}(f_{D,L,s}))\\
        + & (1+\epsilon)^2\left(\ec_{D}(f_{D,L,s})-\ec_D(f_{L_\epsilon}^{w,b})\right)\\
        + & (1+\epsilon)^2\left(\ec_{D}(f_{L_\epsilon}^{w,b})-\ec(f_{L_\epsilon}^{w,b})\right)\\
        + & (1+\epsilon)^2\left(\ec(f_{L_\epsilon}^{w,b})-\ec(f_\rho)\right)\\
        + & \left((1+\epsilon)^2-1\right)\ec(f_\rho)\\
        =: & \sum_{i=1}^8S_i.
    \end{align*}

    We will use an inequality, which we will require through the rest of the steps:
    \begin{equation} \label{ineq:9.1}
        (a+b)^2\leq(1+\epsilon)a^2+(1+1/\epsilon)b^2
    \end{equation} for $a,b,\epsilon>0$. 

    We will bound each of these $S_i$s to prove the universal consistency as done in Part 3, Appendix A in \cite{lin}. 

    We start with $S_1$, we have
    \begin{align*}
        S_1 & = \epsilon(\pi_M(f_{D,L,s}))-(1+\epsilon)(\pi_M(f_{D,L,s}))\\
        & = \int_{\mathcal{Z}}|\pi_{M}(f_{D,L,s}(x))-(\log_0^c(y_M))+(\log_0^c(y_M))-(\log_0^c(y))|^2d\rho\\
        & - (1+\epsilon)\int_{\mathcal{Z}}|\pi_M(f_{D,L,s})-(\log_0^c(y_M))|^2d\rho\\
        & \leq (1+(1/\epsilon))\int_{\mathcal{Z}}|\log_0^c(y)-\log_0^c(y_M)|^2d\rho.
    \end{align*}
    But we have $M=M_m\to\frac{1}{\sqrt{c}}$ as $m\to\infty$. Since $\epsilon>0$ is arbitrary, we get $S_1\to 0$ as $m\to\infty$. 

    By Lemma \ref{lem:19.1} and the constraints in the statement of Theorem \ref{thm:5.1} we get
    \begin{align*}
        S_2\to 0 \hspace{1ex} \text{as} \hspace{1ex} m\to\infty.
    \end{align*}

   By the definition of the truncation operator, we get
   \begin{align*}
       S_3=\frac{1}{m}\sum_{i=1}^m|\pi_M(f_{D,L,s}(x_i))-(\log_0^c(y_{i,M}))|^2-\frac{1}{m}\sum_{i=1}^m|f_{D,L,s}(x_i)-(\log_0^c(y_{i,M}))|^2\leq 0.
   \end{align*}

   By the Strong Law of Large Numbers and inequality \ref{ineq:9.1} we have
   \begin{align*}
       S_4 \leq  (1+\epsilon)(1+1/\epsilon)\frac{1}{m}\sum_{i=1}^m|\log_0^c(y_i)-\log_0^c(y_{i,M})|^2
       \to (1+\epsilon)(1+1/\epsilon)\int_{\zc}|\log_0^c(y)-\log_0^c(y_M)|^2d\rho
   \end{align*}
   as $m\to\infty$ almost surely. By the fact that $M_m\to\frac{1}{\sqrt{c}}$ as $m\to\infty$, we get
   \begin{align*}
       S_4\to 0.
   \end{align*}

   Since $f_{D,L}$ is the HERM estimator, we obtain
   \begin{align*}
       S_5=(1+\epsilon)^2\left(\frac{1}{m}\sum_{i=1}^m|f_{D,L}(x_i)-\log_0^c(y_i)|^2-\frac{1}{m}\sum_{i=1}^m|f_{L_\epsilon}^{w,b}(x_i)-\log_0^c(y_i)|^2\right)\leq 0.
   \end{align*}

    Again by the Strong Law of Large Numbers, we have
    \begin{align*}
        S_6\to 0
    \end{align*}
    almost surely. 

    For $S_7$ we have
    \begin{align*}
        S_7=(1+\epsilon)^2\|f_{L_\epsilon}-f\rho\|^2_{L^2_{\rho_\xx}}.
    \end{align*}
   By Lemma \ref{lem:19.1}, we get
   \begin{align*}
       S_7\leq (1+\epsilon)^2\epsilon.
   \end{align*}

   Also, we have
   \begin{align*}
       S_8\leq ((1+\epsilon)^2-1)\int_{\mathcal{Z}}|f_\rho(x)-\log_0^c(y)|^2d\rho = \epsilon(epsilon+2)\int_{\mathcal{Z}}|f_\rho(x)-\log_0^c(y)|^2d\rho.
   \end{align*}

   Summing up all the terms from $S_1$ to $S_8$, we get
   \begin{align*}
       \limsup_{m\to\infty}\ec(\pi_M(f_{D,L,s}))-\ec(f_\rho)\leq (1+\epsilon)^2\epsilon+\epsilon(2+\epsilon)\int_{\mathcal{Z}}|f_\rho(x)-\log_0^c(y)|^2d\rho
   \end{align*}
   holds almost surely. As $\epsilon>0$ is arbitrary, we can write
   \begin{align*}
       \limsup_{m\to\infty}\ec(\pi_M(f_{D,L,s}))-\ec(f_\rho)=0.
   \end{align*}
   This completes the proof of the universal consistency of eHDCNN. 
\end{proof}



\section*{Computational Resources} 
All experiments were carried out on a personal computer with $12$th Gen Intel(R) Core(TM) i5-1230U 1.00 GHz Processor, 16 GB RAM, Windows $11$ Home $22$H$2$, and Python $3.11.5$.

\bibliographystyle{IEEEtran}
\bibliography{ref}

\clearpage

\end{document}